\documentclass{article}

\usepackage{graphicx} 
\usepackage{subfigure}

\usepackage[accepted]{icml2017}
\usepackage{url}
\usepackage[all]{xy}
\usepackage{color}
\usepackage{rotating}
\usepackage{amsfonts}
\usepackage{amsmath}
\usepackage{bm}
\usepackage{wrapfig}
\usepackage{times}
\usepackage{natbib}
\icmltitlerunning{The Predictron: End-To-End Learning and Planning}

\newcommand{\given}{\ | \ }

\newcommand{\expectx}[2]{\mathbb{E}_{#1} \left[ #2 \right]}

\newcommand{\dd}[2]{\frac{\partial{#1}}{\partial{#2}}}

\newcommand{\blue}[1]{{\color{blue}{#1}}}
\newcommand{\red}[1]{{\color{red}{#1}}}

\newcommand{\orange}[1]{{\color{magenta}{#1}}}
\newcommand{\cyan}[1]{{\textcolor{cyan}{#1}}}

\definecolor{mygreen}{RGB}{0, 140, 80}
\newcommand{\mygreen}[1]{{\color{mygreen}{#1}}}
\definecolor{darkgrey}{RGB}{30, 30, 30}
\newcommand{\darkgrey}[1]{{\color{darkgrey}{#1}}}

\newcommand{\gR}{\mathbf{g}}
\newcommand{\rR}{\mathbf{r}}
\newcommand{\dR}{\pmb{\gamma}}

\newcommand{\g}{\mathbf{g}}
\renewcommand{\r}{\mathbf{r}}
\renewcommand{\d}{\pmb{\gamma}}
\newcommand{\s}{\mathbf{s}}
\renewcommand{\v}{\mathbf{v}}
\newcommand{\I}{\bm{1}}

\newcommand{\w}{\mathbf{w}}
\renewcommand{\l}{\pmb{\lambda}}

\newcommand{\bl}{\bm{w}}

\newcommand{\zero}{\bm{0}}
\renewcommand{\th}{\bm{\theta}}
\newcommand{\et}{\bm{\eta}}

\begin{document} 

\twocolumn[
\icmltitle{The Predictron: End-To-End Learning and Planning}

 \icmlsetsymbol{equal}{*}
\begin{icmlauthorlist}
\icmlauthor{David Silver}{equal,dm}
\icmlauthor{Hado van Hasselt}{equal,dm}
\icmlauthor{Matteo Hessel}{equal,dm}
\icmlauthor{Tom Schaul}{equal,dm}
\icmlauthor{Arthur Guez}{equal,dm}
\icmlauthor{Tim Harley}{dm}
\icmlauthor{Gabriel Dulac-Arnold}{dm}
\icmlauthor{David Reichert}{dm}
\icmlauthor{Neil Rabinowitz}{dm}
\icmlauthor{Andre Barreto}{dm}
\icmlauthor{Thomas Degris}{dm}
\icmlaffiliation{dm}{DeepMind, London}
\end{icmlauthorlist}

\icmlcorrespondingauthor{David Silver}{davidsilver@google.com}
\icmlcorrespondingauthor{Hado van Hasselt}{hado@google.com}
\icmlcorrespondingauthor{Matteo Hessel}{mtthss@google.com}
\icmlcorrespondingauthor{Tom Schaul}{schaul@google.com}
\icmlcorrespondingauthor{Arthur Guez}{aguez@google.com}

\icmlkeywords{reinforcement learning, prediction, planning, deep learning, abstract models}

\vskip 0.3in
]
\printAffiliationsAndNotice{\icmlEqualContribution}

\begin{abstract} 
One of the key challenges of artificial intelligence is to learn models that are effective in the context of planning. In this document we introduce the \emph{predictron} architecture. The predictron consists of a fully abstract model, represented by a Markov reward process, that can be rolled forward multiple ``imagined" planning steps. Each forward pass of the predictron accumulates internal rewards and values over multiple planning depths. 
The predictron is trained end-to-end so as to make these accumulated values accurately approximate the true value function.
We applied the predictron to procedurally generated random mazes and a simulator for the game of pool. The predictron yielded significantly more accurate predictions than conventional deep neural network architectures.
\end{abstract} 

\section{Introduction}

The central idea of model-based reinforcement learning is to decompose the RL problem into two subproblems: learning a model of the environment, and then planning with this model. The model is typically represented by a Markov reward process (MRP) or decision process (MDP). The planning component uses this model to evaluate and select among possible strategies. This is typically achieved by rolling forward the model to construct a value function that estimates cumulative reward. In prior work, the model is trained essentially independently of its use within the planner. As a result, the model is not well-matched with the overall objective of the agent. Prior deep reinforcement learning methods have successfully constructed models that can unroll near pixel-perfect reconstructions \citep{Oh:2015,Chiappa:2016}; but are yet to surpass state-of-the-art model-free methods in challenging RL domains with raw inputs \citep[e.g.,][]{Mnih:2015,Mnih:2016,Lillicrap:2016}.

In this paper we introduce a new architecture, which we call the \emph{predictron}, that integrates learning and planning into one end-to-end training procedure. At every step, a model is applied to an internal state, to produce a next state, reward, discount, and value estimate. This model is completely abstract and its only goal is to facilitate accurate value prediction.
For example, to plan effectively in a game, an agent must be able to predict the score. If our model makes accurate predictions, then an optimal plan with respect to our model will also be optimal for the underlying game -- even if the model uses a different state space (e.g., abstract representations of enemy positions, ignoring their shapes and colours), action space (e.g., high-level actions to move away from an enemy), rewards (e.g., a single abstract step could have a higher value than any real reward), or even time-step (e.g., a single abstract step could ``jump" the agent to the end of a corridor).
All we require is that trajectories through the abstract model produce scores that are consistent with trajectories through the real environment. This is achieved by training the predictron end-to-end, so as to make its value estimates as accurate as possible. 

An ideal model could generalise to many different prediction tasks, rather than overfitting to a single task; and could learn from a rich variety of feedback signals, not just a single extrinsic reward. We therefore train the predictron to predict a host of different value functions for a variety of pseudo-reward functions and discount factors. These pseudo-rewards can encode any event or aspect of the environment that the agent may care about, e.g., staying alive or reaching the next room.

We focus upon the prediction task: estimating value functions in MRP environments with uncontrolled dynamics.
In this case, the predictron can be implemented as a deep neural network with an MRP as a recurrent core. The predictron unrolls this core multiple steps and accumulates rewards into an overall estimate of value. 

We applied the predictron to procedurally generated random mazes, and a simulated \emph{pool} domain, directly from pixel inputs. In both cases, the predictron significantly outperformed model-free algorithms with conventional deep network architectures; and was much more robust to architectural choices such as depth.

\section{Background}

We consider environments defined by an MRP with states $s \in \mathcal{S}$. The MRP is defined by a function, $s', r, \gamma = p(s, \alpha)$, where $s'$ is the next state, $r$ is the reward, and $\gamma$ is the discount factor, which can for instance represent the non-termination probability for this transition. The process may be stochastic, given IID noise $\alpha$. 

The \emph{return} of an MRP is the cumulative discounted reward over a single trajectory, $g_t = r_{t+1} + \gamma_{t+1} r_{t+2} + \gamma_{t+1} \gamma_{t+2} r_{t+3} + ...\,$, where $\gamma_t$ 
can vary per time-step.
We consider a generalisation of the MRP setting that includes vector-valued rewards $\rR$, diagonal-matrix discounts $\dR$, and vector-valued returns $\gR$; definitions are otherwise identical to the above. We use this bold font notation to closely match the more familiar scalar MRP case; the majority of the paper can be comfortably understood by reading all rewards as scalars, and all discount factors as scalar and constant, i.e., $\gamma_t = \gamma$.

The \emph{value function} of an MRP $p$ is the expected return from state $s$, $v_p(s) = \expectx{p}{\gR_t \given s_t = s}$. In the vector case, these are known as \emph{general} value functions \citep{Sutton:2011}.
We will say that a (general) value function $v(\cdot)$ is \emph{consistent} with environment $p$ if and only if $v = v_p$ which satisfies the following \emph{Bellman equation} \citep{Bellman:1957},
\begin{align}
\label{bellman_environment}
v_p(s) &= \expectx{p}{\rR + \dR v_p(s') \given s} \,.
\end{align}

In model-based reinforcement learning \citep{Sutton:1998book}, an approximation $m \approx p$ to the environment is learned. In the uncontrolled setting this model is normally an MRP $s', \rR, \dR = m(s, \beta)$ that maps from state $s$ to subsequent state $s'$ and additionally outputs rewards $\rR$ and discounts $\dR$; the model may be stochastic given an IID source of noise $\beta$. A (general) value function $v_m(\cdot)$ is consistent with model $m$ (or \emph{valid}, \citep{sutton:mixtures}), if and only if it satisfies a Bellman equation $v_m(s) = \expectx{m}{\rR + \dR v_m(s') \given s}$ with respect to model $m$. Conventionally, model-based RL methods focus on finding a value function $v$ that is consistent with a separately learned model $m$.

\section{Predictron architecture} \label{sec:arch}

\begin{figure*}
\centerline{
\begin{xy}
\centering
	\xymatrix@R-1.5pc@C-0.5pc{
	     \ar@{|.|}[rrrrrrrr]|{\text{\;\; a) $k$-step predictron \;\;}} &&& &&& &&& 
	     \ar@{|.|}[rr]|{\;\;\text{b) $\lambda$-predictron \;\;}} && \\
	     &&&
	     &&&
	     \rotatebox{90}{...} &&&
	     \rotatebox{90}{...} && \rotatebox{90}{...} \ar[dd]^{\darkgrey{\gamma^2}\mygreen{\lambda^2}} \\
	     &&&
	     &&&
	     &&&
	     & \red{\r^2} \ar[dr]^{\mygreen{\lambda^2}} & \\
	     &&&
	     \rotatebox{90}{...} &&&
		 \cyan{\s^2} \ar[r] \ar@[lightgray][uu] \ar@[lightgray][ur]& \blue{\v^2} \ar[r] & + \ar[dd]^{\darkgrey{\gamma^1}} &
		 \cyan{\s^2} \ar[uu] \ar[r] \ar[ur] & \blue{\v^2} \ar[r]_{\mygreen{1-\lambda^2}} & \ar[dd]^{\darkgrey{\gamma^1}\mygreen{\lambda^1}} + \\ 
		 &&&
		 &&&
		 & \red{\r^1} \ar[dr] &&
		 & \red{\r^1} \ar[dr]^{\mygreen{\lambda^1}} && \\
		 \rotatebox{90}{...} &&& 
		 \cyan{\s^1} \ar[r] \ar@[lightgray][uu] \ar@[lightgray][ur] & \blue{\v^1} \ar[r] & + \ar[dd]^{\darkgrey{\gamma^0}} &
		 \cyan{\s^1} \ar[uu] \ar@[lightgray][r] \ar[ur] &  & + \ar[dd]^{\darkgrey{\gamma^0}} &
		 \cyan{\s^1} \ar[uu] \ar[r] \ar[ur] & \blue{\v^1} \ar[r]_{\mygreen{1-\lambda^1}} & \ar[dd]^{\darkgrey{\gamma^0}\mygreen{\lambda^0}}  + \\
		 &&&
		 & \red{\r^0} \ar[dr] && 
		 & \red{\r^0} \ar[dr] &&
		 & \red{\r^0} \ar[dr]^{\mygreen{\lambda^0}} & \\
		 \cyan{\s^0} \ar[r] \ar@[lightgray][uu] \ar@[lightgray][ur] & \blue{\v^0} \ar[r] & + \ar[dd] &
		 \cyan{\s^0} \ar[uu] \ar@[lightgray][r] \ar[ur] &  & + \ar[dd] & 
		 \cyan{\s^0} \ar[uu] \ar@[lightgray][r] \ar[ur] &  & + \ar[dd] &
         \cyan{\s^0} \ar[uu] \ar[r] \ar[ur] & \blue{\v^0} \ar[r]_{\mygreen{1-\lambda^0}} & \ar[dd]^{\ } + \\ 
		 \\
		 s \ar[uu] && \orange{\g^0} &
		 s \ar[uu] && \orange{\g^1} & 
		 s \ar[uu] && \orange{\g^2} &
		 s \ar[uu] && \orange{\g^\lambda} \\
    }
\end{xy}
}
\caption{
a) The $k$-step predictron architecture. The first three columns illustrate 0, 1 and 2-step pathways through the predictron. The 0-step preturn reduces to standard model-free value function approximation; other preturns ``imagine" additional steps with an internal model. Each pathway outputs a $k$-step preturn $\g^k$ that accumulates discounted rewards along with a final value estimate. In practice all $k$-step preturns are computed in a single forward pass. b) The $\lambda$-predictron architecture. The $\lambda$-parameters gate between the different preturns. The output is a $\lambda$-preturn $\g^{\lambda}$ that is a mixture over the $k$-step preturns. For example, if $\l^0=\I, \l^1=\I, \l^2=\zero$ then we recover the 2-step preturn, $\g^\lambda=\g^2$. Discount factors $\d^k$ and $\lambda$-parameters $\l^k$ are dependent on state $\s^k$; this dependence is not shown in the figure.
}
\label{fig:architecture}
\end{figure*}
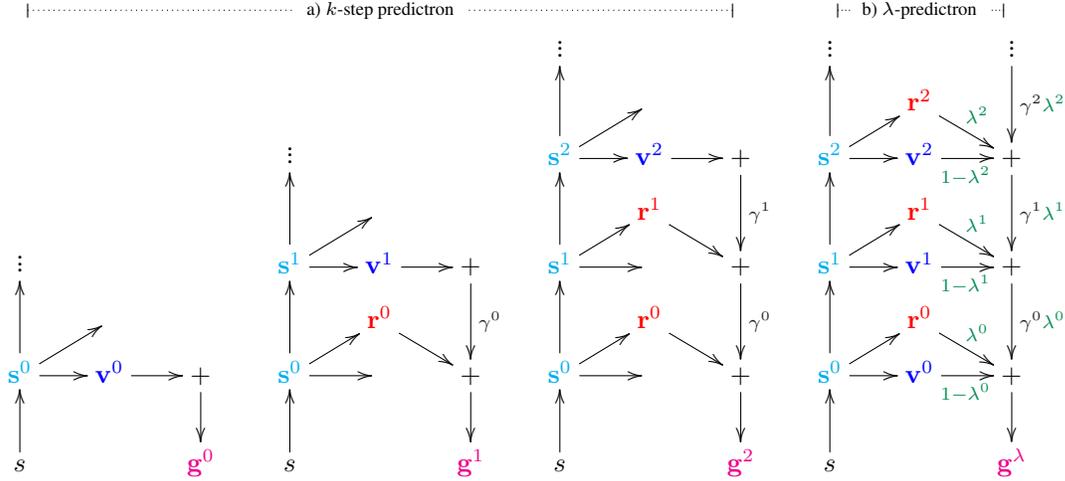

The predictron is composed of four main components. First, a state representation $\s = f(s)$ that encodes raw input $s$ (this could be a history of observations, in partially observed settings, for example when $f$ is a recurrent network) into an internal (abstract, hidden) state $\s$. Second, a model $\s', \r, \d = m(\s, \beta)$ that maps from internal state $\s$ to subsequent internal state $\s'$, internal rewards $\r$, and internal discounts $\d$.
Third, a value function $v$ that outputs internal values $\v = v(\s)$ representing the remaining internal return from internal state $\s$ onwards. The predictron is applied by unrolling its model $m$ multiple ``planning" steps to produce internal rewards, discounts and values. We use superscripts $\bullet^k$ to indicate internal steps of the model (which have no necessary connection to time steps $\bullet_t$ of the environment). 
Finally, these internal rewards, discounts and values are combined together by an \emph{accumulator} into an overall estimate of value $\g$. The whole predictron, from input state $s$ to output, may be viewed as a value function approximator for external targets (i.e., the returns in the real environment). We consider both $k$-step and $\lambda$-weighted accumulators.

The \emph{$k$-step predictron} rolls its internal model forward $k$ steps (Figure \ref{fig:architecture}a). The 0-step \emph{predictron return} (henceforth abbreviated as \emph{preturn}) is simply the first value $\g^0 = \v^0$, the 1-step preturn is $\g^1 = \r^1 + \d^1 \v^1$. More generally, the $k$-step \emph{predictron return} $\g^k$  is the internal return obtained by accumulating $k$ model steps, plus a discounted final value $\v^k$ from the $k$th step:

\begin{align*}
\g^k  = \r^1 + \d^1 (\r^2 + \d^2 ( \ldots + \d^{k-1}(\r^k + \d^k \v^k) \ldots))
\end{align*}

The \emph{$\lambda$-predictron} combines together many $k$-step preturns.
Specifically, it computes a diagonal weight matrix $\l^k$ from each internal state $s^k$. 
The accumulator uses weights $\l^0, ..., \l^K$ to aggregate over $k$-step preturns $\g^0, ..., \g^K$ and output a combined value that we call the \emph{$\lambda$-preturn} $\g^{\lambda}$,
\begin{align}
\label{lambda_preturn}
\g^{\lambda} &= \sum_{k=0}^K \bl^k \g^k
\end{align}
\begin{align}
\label{lambda_preturn_weights}
\bl^k = \begin{cases}
 (\I - \l^k) \prod_{j=0}^{k-1} \l^j\, & \text{if}\; k < K \\
 \\
 \prod_{j=0}^{K-1} \l^j & \text{otherwise.}
 \end{cases}
\end{align}
where $\I$ is the identity matrix. This $\lambda$-preturn is analogous to the $\lambda$-return in the forward-view TD($\lambda$) algorithm \citep{Sutton:1988,Sutton:1998book}. It may also be computed by a backward accumulation through intermediate steps $\g^{k,\lambda}$,
\begin{align}
\g^{k,\lambda} &= (\I -\l^k) \v^k + \l^k \left( \r^{k+1} + \d^{k+1} \g^{k+1,\lambda} \right) \,,
\label{eqn:lambda-preturn}
\end{align}
where $\g^{K, \lambda} = \v^K$, and then using $\g^\lambda = \g^{0, \lambda}$.
Computation in the $\lambda$-predictron operates in a sweep, iterating first through the model from $k=0\ldots K$ and then back through the accumulator from $k=K\ldots 0$ in a single ``forward" pass of the network (see Figure \ref{fig:architecture}b). 
Each $\l^k$ weight acts as a gate on the computation of the $\lambda$-preturn: a value of $\l^k = \zero$ will truncate the $\lambda$-preturn at layer $k$, while a value of $\l^k = \I$ will utilise deeper layers based on additional steps of the model $m$; the final weight is always $\l^K=\zero$. The individual $\l^k$ weights may depend on the corresponding abstract state $\s^k$ and can differ per prediction. This enables the predictron to compute to an adaptive depth \citep{graves:act} depending on the internal state and learning dynamics of the network. 

\section{Predictron learning updates}
\label{sec:updates}

We first consider updates that optimise the joint parameters $\th$ of the state representation, model, and value function. 
We begin with the $k$-step predictron. We update the $k$-step preturn $\g^k$ towards a target outcome $\g$, e.g. the Monte-Carlo return from the real environment, by minimising a mean-squared error loss,
\begin{align}
 & L^k = \frac{1}{2} \left\| \expectx{p}{\g \given s} - \expectx{m}{\g^{k} \given s} \right\|^2 \,.\notag\\
& \dd{l^k}{\th} = \left( \g - \g^{k} \right) \dd{\g^{k}}{\th} \,.
\label{eqn:k-loss}
\end{align}
where $l^k = \frac{1}{2} \left\| \g - \g^{k} \right\|^2$ is the sample loss. We can use the gradient of the sample loss to update parameters, e.g., by stochastic gradient descent. 
For stochastic models, independent samples of $\g^{k}$ and $\dd{\g^{k}}{\th}$ are required for unbiased samples of the gradient of $L^k$.

The $\lambda$-predictron combines many $k$-step preturns. To update the joint parameters $\th$, we can uniformly average the losses on the individual preturns $\g^k$,
\begin{align}
&  L^{0:K} = \frac{1}{2K}\sum_{k=0}^K
 \left\| \expectx{p}{\g \given s} - \expectx{m}{\g^{k} \given s} \right\|^2 \,,\notag\\
& \dd{l^{0:K}}{\th} = \frac{1}{K} \sum_{k=0}^K
\left( \g - \g^{k} \right) \dd{\g^{k}}{\th} \,.
\label{eqn:multi-k-loss}
 \end{align}
Alternatively, we could weight each loss by the usage $\bl^k$ of the corresponding preturn, such that the gradient is $\sum_{k=0}^K \bl^k
\left( \g - \g^{k} \right) \dd{\g^{k}}{\th}$.

In the $\lambda$-predictron, the $\l^k$ weights (that determine the relative weighting $\bl^k$ of the $k$-step preturns) depend on additional parameters $\et$, which are updated so as to minimise a mean-squared error loss $L^\lambda$,
\begin{align}
& L^\lambda = \frac{1}{2}\left\| \expectx{p}{\g  \given  s} - \expectx{m}{\g^\lambda  \given  s} \right\|^2 \,.\notag\\
& \dd{l^\lambda}{\et} = \left( \g - \g^\lambda \right) \dd{\g^\lambda}{\et}.
\label{eqn:lambda-update}
\end{align}

In summary, the joint parameters $\th$ of the state representation $f$, the model $m$, and the value function $v$ are updated to make each of the $k$-step preturns $\g^k$ more similar to the target $\g$, and the parameters $\et$ of the $\lambda$-accumulator are updated to learn the weights $\bl^k$ so that the aggregate $\lambda$-preturn $\g^{\lambda}$ becomes more similar to the target $\g$.

\subsection{Consistency updates}
\label{sec:semi}

In model-based reinforcement learning architectures such as Dyna \citep{Sutton:1990dyna}, value functions may be updated using both real and imagined trajectories. The refinement of value estimates based on these imagined trajectories is often referred to as \textit{planning}. A similar opportunity arises in the context of the predictron. Each rollout of the predictron generates a trajectory in abstract space, alongside with rewards, discounts and values. Furthermore, the predictron aggregates these components in multiple value estimates ($\g^{0}$, ..., $\g^{k}$, $\g^\lambda$). 

We may therefore update each individual value estimate towards the best aggregated estimate. This corresponds to adjusting each preturn $\g^k$ towards the $\lambda$-preturn $\g^\lambda$, by minimizing:
\begin{align}
& L = \frac{1}{2}\sum_{k=0}^K \left\| \expectx{m}{\g^\lambda \given s} - \expectx{m}{\g^{k} \given s} \right\|^2 \,.\notag\\
& \dd{l}{\th} = \sum_{k=0}^K \left( \g^\lambda - \g^{k} \right) \dd{\g^{k}}{\th}\,.
\label{eqn:consistency}
\end{align}
Here $\g^\lambda$ is considered fixed; the parameters $\th$ are only updated to make $\g^k$ more similar to $\g^\lambda$, not vice versa.

These consistency updates do not require any labels $\g$ or samples from the environment. As a result, it can be applied to (potentially hypothetical) states that have no associated `real' (e.g. Monte-Carlo) outcome: we update the value estimates to be self-consistent with each other. This is especially relevant in the semi-supervised setting, where these consistency updates allow us to exploit the unlabelled inputs.

\section{Experiments}

We conducted experiments in two domains. The first domain consists of randomly generated mazes. Each location either is empty or contains a wall. In these mazes, we considered two tasks. In the first task, the input was a $13\times 13$ maze and a random initial position and the goal is to predict a trajectory generated by a simple fixed deterministic policy. The target $\g$ was a vector with an element for each cell of the maze which is either one, if that cell was reached by the policy, or zero.
In the second random-maze task the goal was to predict for each of the cells on the diagonal of a $20\times 20$ maze (top-left to bottom-right) whether it is connected to the bottom-right corner. Two locations in a maze are considered connected if they are both empty and we can reach one from the other by moving horizontally or vertically through adjacent empty cells. In both cases some predictions would seem to be easier if we could learn a simple algorithm, such as some form of search or flood fill; our hypothesis is that an internal model can learn to emulate such algorithms, where naive approximation may struggle. A few example mazes are shown in Figure 2.

\begin{figure}
\centering
\includegraphics[width=0.27\textwidth]{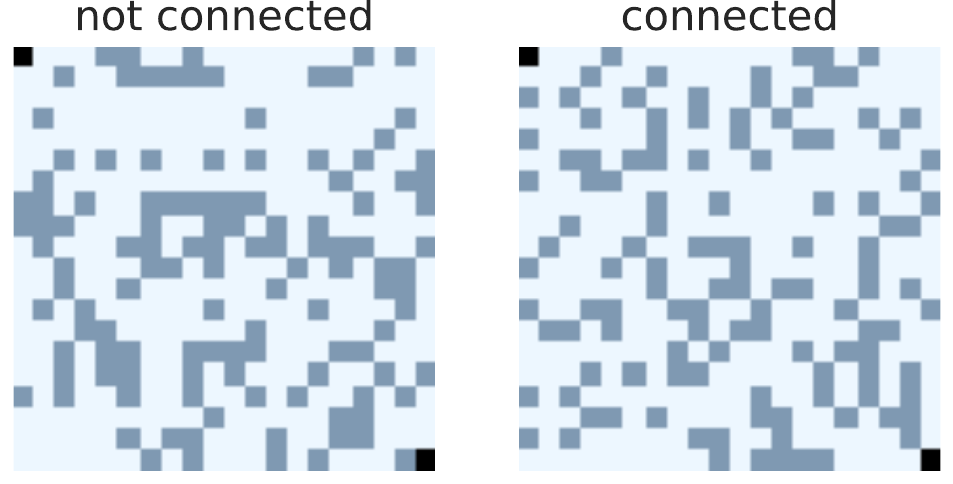}\\
\includegraphics[width=0.47\textwidth]{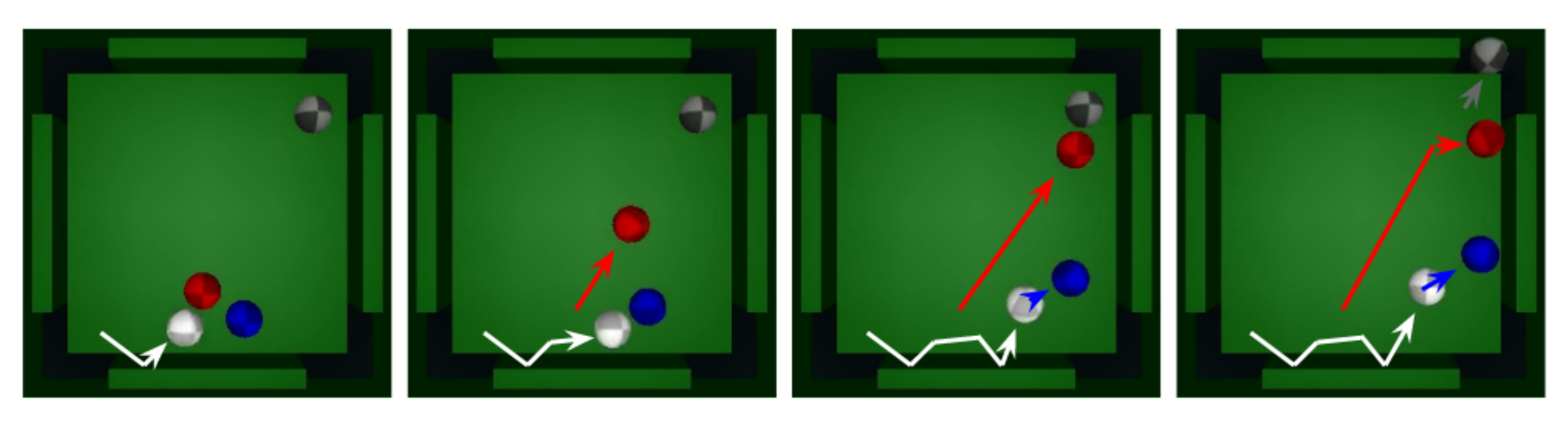}
\vspace{-0.2cm}
\caption{\textbf{Top:} Two sample mazes from the random-maze domain.  Light blue cells are empty, darker blue cells contain a wall. One maze is connected from top-left to bottom-right, the other is not. \textbf{Bottom:} An example trajectory in the pool domain (before downsampling), selected by maximising the prediction by a predictron of pocketing balls. \label{fig:domains}}
\vspace{-0.2cm}
\end{figure}

Our second domain is a simulation of the game of pool, using four balls and four pockets. The simulator is implemented in the physics engine \emph{Mujoco} \citep{Todorov:2012}. We generate sequences of RGB frames starting from a random arrangement of balls on the table. 
The goal is to simultaneously learn to predict future events for each of the four balls, given 5 RGB frames as input. These events include: collision with any other ball, collision with any boundary of the table, entering a quadrant ($\times$4, for each quadrant), being located in a quadrant ($\times$4, for each quadrant), and entering a pocket ($\times$4, for each pocket). Each of these $14 \times 4$ events provides a binary pseudo-reward that we combine with 5 different discount factors $\{0, 0.5, 0.9, 0.98, 1\}$ and predict their cumulative discounted sum over various time spans. This yields a total of 280 general value functions. An example trajectory is shown in Figure \ref{fig:domains}. 
In both domains, inputs are presented as minibatches of i.i.d.~samples with their regression targets.
Additional domain details are provided in the appendix.

\subsection{Learning sequential plans}
\label{sec:seqplans}

\begin{figure}[t]
\centering
\includegraphics[width=0.48\textwidth]{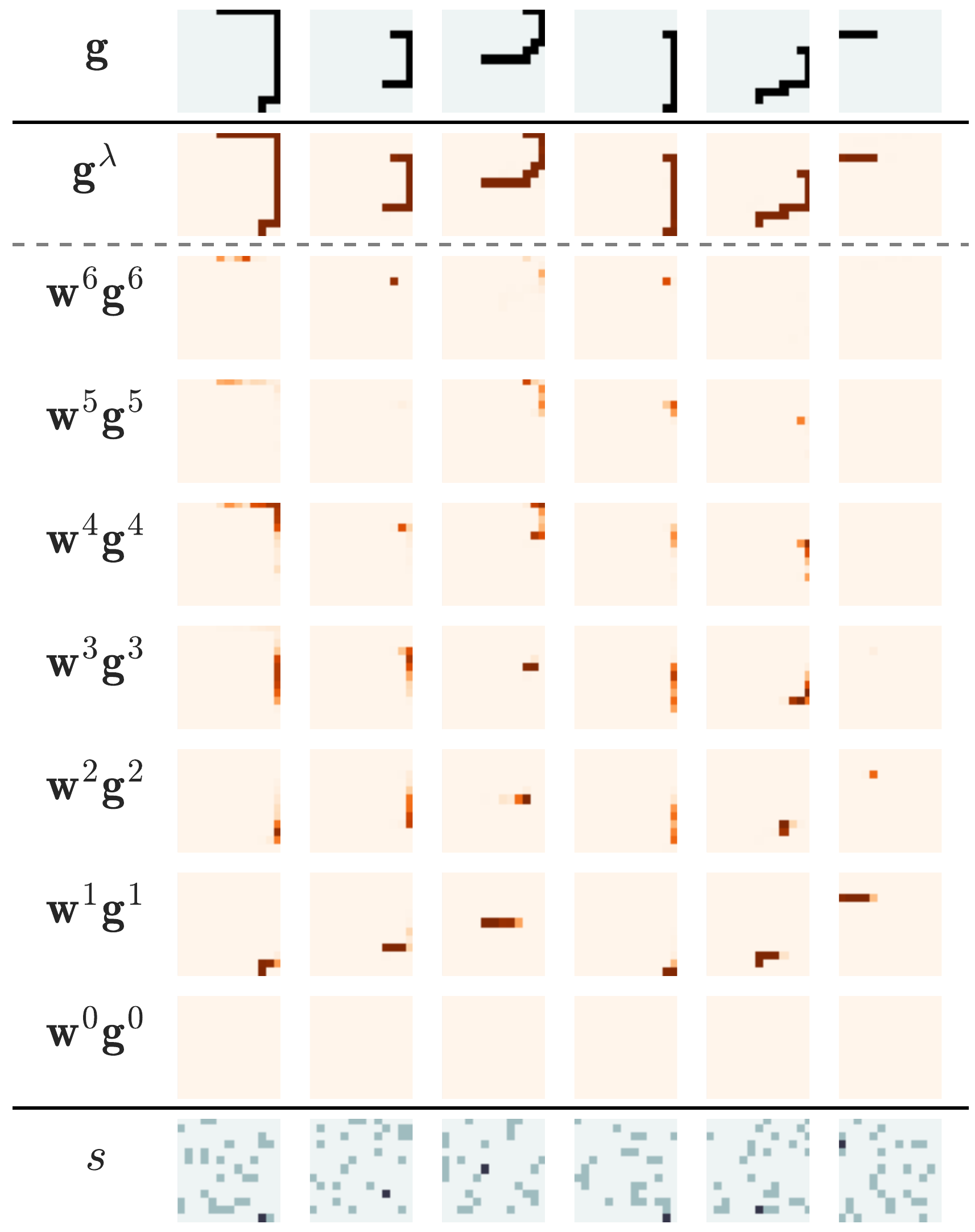}
\vspace{-0.2cm}
\caption{\textbf{Indication of planning.} Sampled mazes (grey) and start positions (black) are shown superimposed on each other at the bottom.  The corresponding target vector $\g$, arranged as a matrix for visual clarity, is shown at the top.  The ensembled prediction $\sum_k \w^k \g^k = \g^{\l}$ is shown just below the target---the prediction is near perfect.  The weighted preturns $\w^k \g^k$ that make up the prediction are shown below $\g^{\l}$. We can see that full predicted trajectory is built up in steps, starting at the start position and then planning through the trajectory in sequence. \label{predictron_planning}}
\vspace{-0.2cm}
\end{figure}

In the first experiment we trained a predictron to predict trajectories generated by a simple deterministic policy in $13\times 13$ random mazes with random starting positions. Figure \ref{predictron_planning} shows the weighted preturns $\w^k \g^k$ and the resulting prediction $\g^{\l} = \sum_k \w^k \g^k$ for six example inputs and targets.  The predictions are almost perfect---the training error was very close to zero. The full prediction is composed from weighted preturns which decompose the trajectory piece by piece, starting at the start position in the first step $k=1$, and where often multiple policy steps are added per planning step.  The predictron was not informed about the sequential build up of the targets---it never sees a policy walking through the maze, only the resulting trajectories---and yet sequential plans emerged spontaneously.  Notice also that the easier trajectory on the right was predicted in only two steps, while more thinking steps are used for more complex trajectories.

\subsection{Exploring the predictron architecture}
\begin{figure*}
\centering
\raisebox{.4\height}{\includegraphics[width=0.24\textwidth]{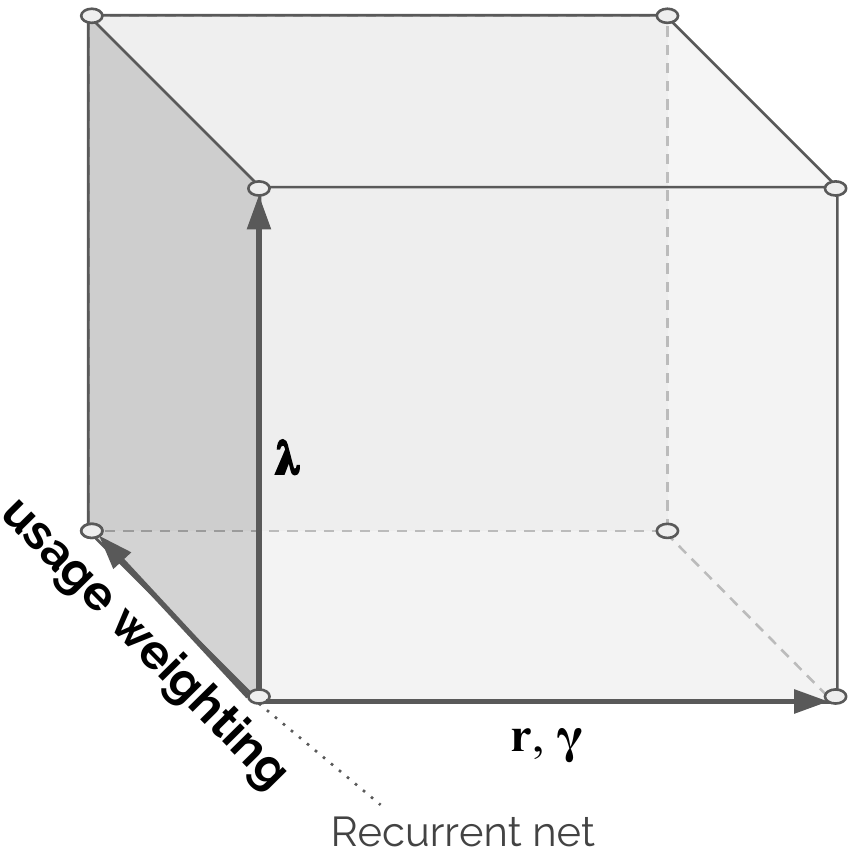}}\hspace{1cm}\includegraphics[width=0.65\textwidth]{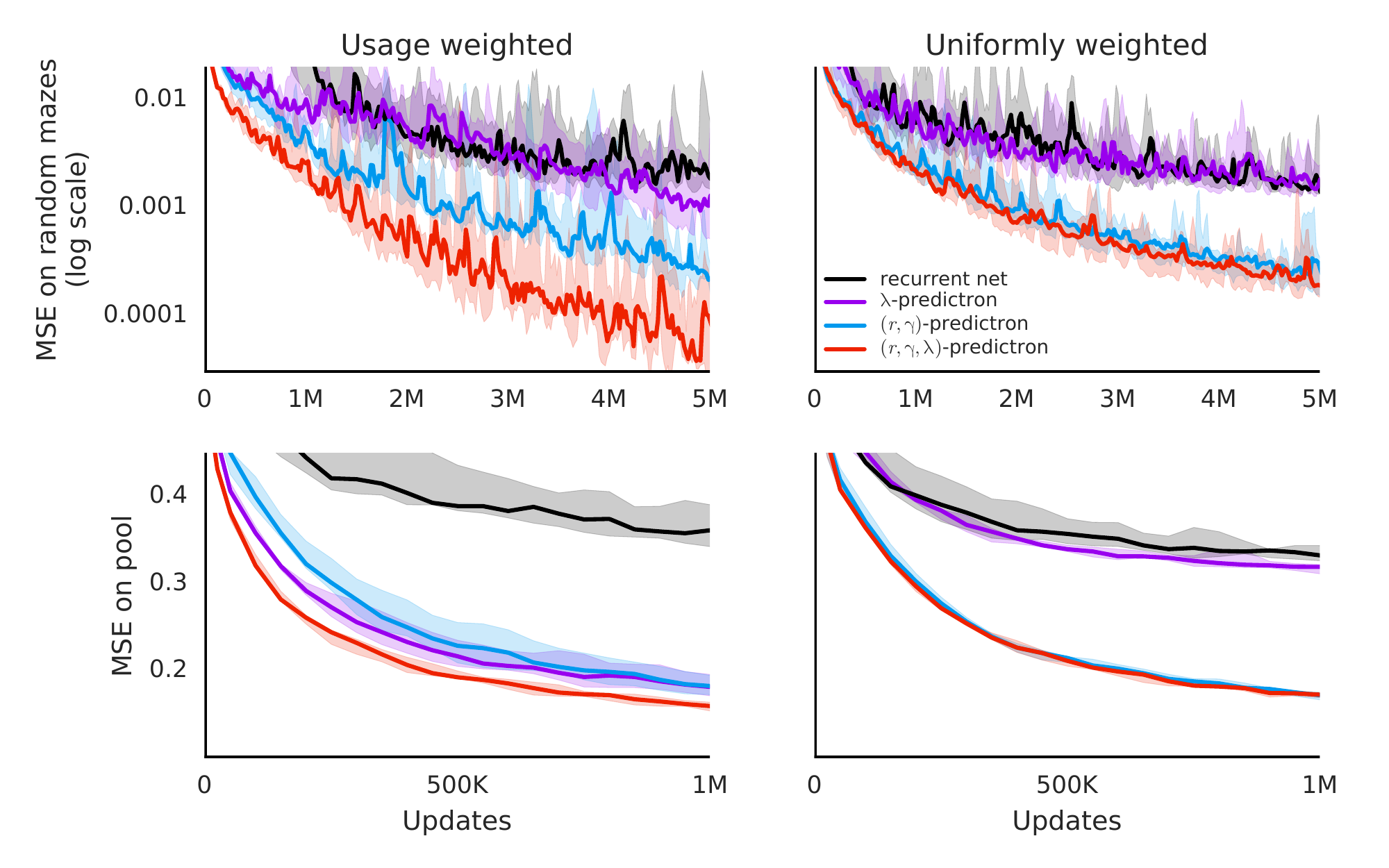}
\vspace{-0.2cm}
\caption{\textbf{Exploring predictron variants.} Aggregated prediction errors over all predictions (20 for mazes, 280 for pool) for the eight predictron variants corresponding to the cube on the left (as described in the main text), for both random mazes (top) and pool (bottom). Each line is the median of RMSE over five seeds; shaded regions encompass all seeds. The full $(r, \gamma, \lambda)$-prediction (\red{\textbf{red}}) consistently performed best.\label{predictron_results_first_cube}}
\vspace{-0.2cm}
\end{figure*}

In the next set of experiments, we tackle the problem of predicting connectivity of multiple pairs of locations in a random maze, and the problem of learning many different value functions from our simulator of the game of pool. We use these more challenging domains to examine three binary dimensions that differentiate the predictron from standard deep networks. We compare eight predictron variants corresponding to the corners of the cube on the left in Figure \ref{predictron_results_first_cube}.

The first dimension, labelled $r,\gamma$, corresponds to whether or not we use the structure of an MRP model. In the MRP case internal rewards and discounts are both learned. In the non-($r,\gamma$) case, which corresponds to a vanilla hidden-to-hidden neural network module, internal rewards and discounts are ignored by fixing their values to $\r^k = {\bm 0}$ and $\d^k = {\bm 1}$. 

The second dimension is whether a $K$-step accumulator or $\lambda$-accumulator is used to aggregate preturns. When a $\lambda$-accumulator is used, a $\lambda$-preturn is computed as described in Section \ref{sec:arch}. Otherwise, intermediate preturns are ignored by fixing $\l^k = 1$ for $k<K$. In this case, the overall output of the predictron is the maximum-depth preturn $\g^K$.

The third dimension, labelled \textit{usage weighting}, defines the loss that is used to update the parameters $\th$. We consider two options: the preturn losses can either be weighted uniformly (see Equation~\ref{eqn:multi-k-loss}), or the update for each preturn $\g^k$ can be weighted according to the weight $\bl^k$ that determines how much it is used in the $\lambda$-predictron's overall output. We call the latter loss `usage weighted'. Note that for architectures without a $\lambda$-accumulator, $\bl^k=0$ for $k<K$, and $\bl^K=1$, thus usage weighting then implies backpropagating only the loss on the final preturn $\g^K$.

All variants utilise a convolutional core with 2 intermediate hidden layers; parameters were updated by supervised learning (see appendix for more details). Root mean squared prediction errors for each architecture, aggregated over all predictions, are shown in Figure \ref{predictron_results_first_cube}. The top row corresponds to the random mazes and the bottom row to the pool domain.
The main conclusion is that learning an MRP model improved performance greatly.  The inclusion of $\l$ weights helped as well, especially on pool. Usage weighting further improved performance.

\subsection{Comparing to other architecture}

\begin{figure*}
\centering
\raisebox{.3\height}{\includegraphics[width=0.26\textwidth]{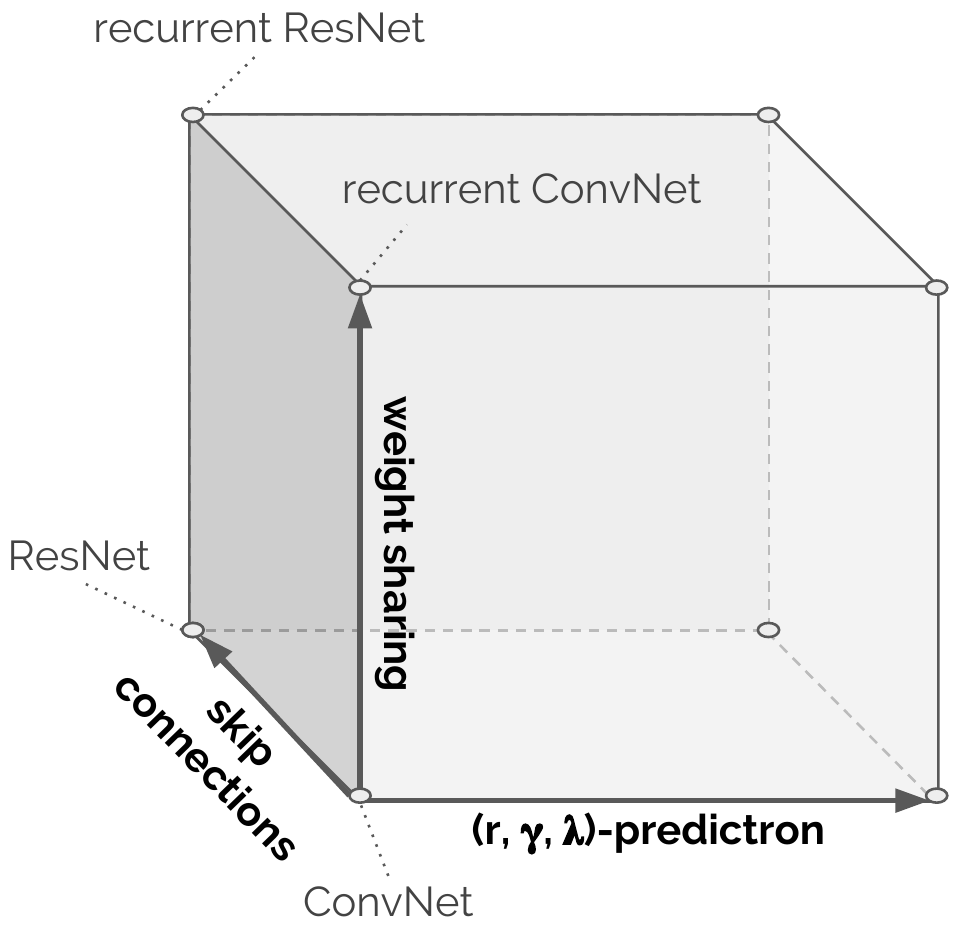}}\hspace{1cm}\includegraphics[width=0.65\textwidth]{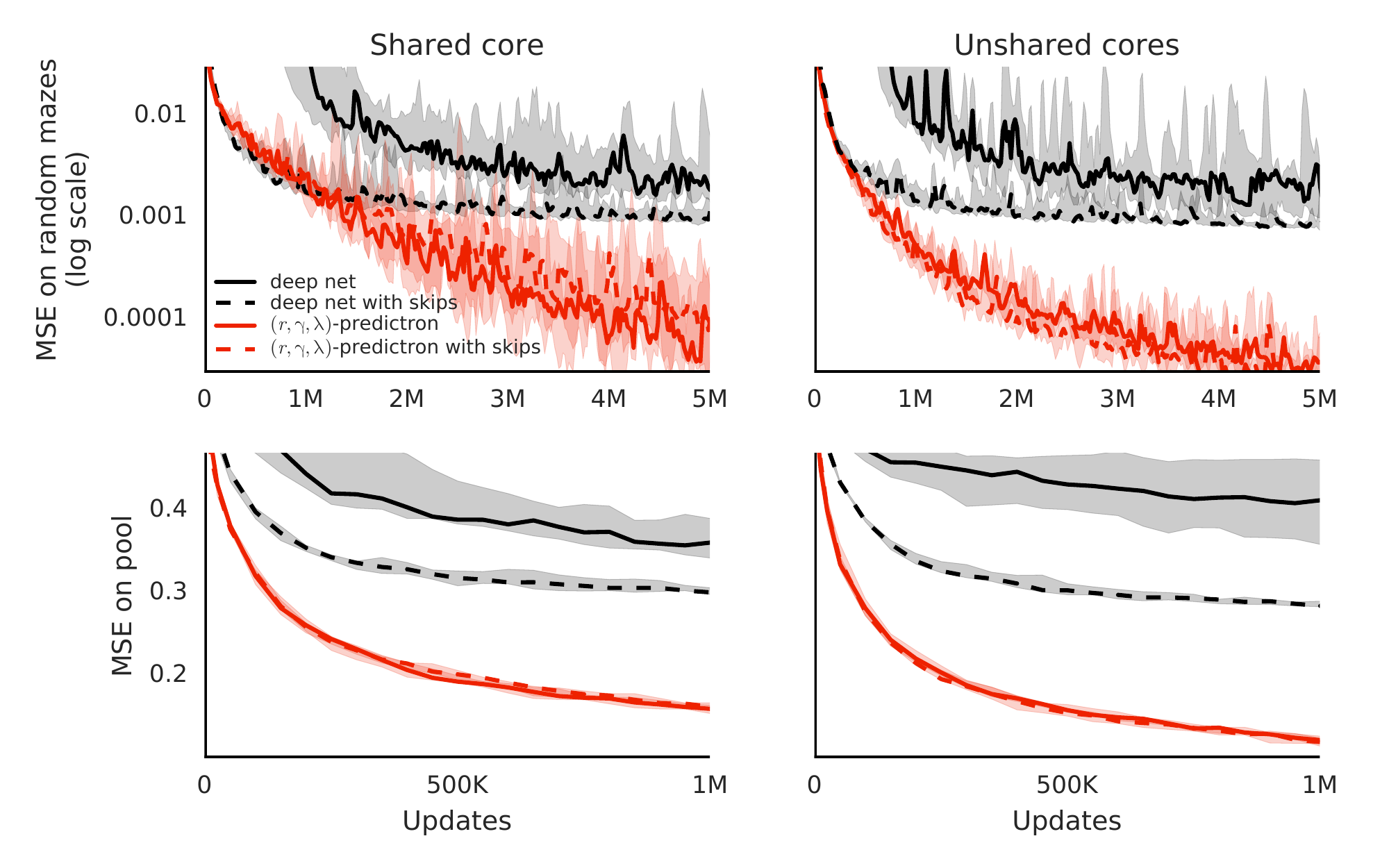}
\vspace{-0.5cm}
\caption{\textbf{Comparing predictron to baselines.} Aggregated prediction errors on random mazes (top) and pool (bottom) over all predictions for the eight architectures corresponding to the cube on the left. Each line is the median of RMSE over five seeds; shaded regions encompass all seeds. The full $(r, \gamma, \lambda)$-predictron (\red{\textbf{red}}), consistently outperformed conventional deep network architectures (\textbf{black}), with and without skips and with and without weight sharing.\label{predictron_results_second_cube}}
\vspace{-0.2cm}
\end{figure*}

Our third set of experiments compares the predictron to feedforward and recurrent deep learning architectures, with and without skip connections. We compare the corners of a new cube, as depicted on the left in Figure \ref{predictron_results_second_cube}, based on three different binary dimensions.

The first dimension of this second cube is whether we use a predictron, or a (non-$\lambda$, non-$(r,\gamma)$) deep network that does not have an internal model and does not output or learn from intermediate predictions. We use the most effective predictron from the previous section, i.e., the $(r,\gamma,\lambda)$-predictron with usage weighting. 

The second dimension is whether all cores share weights (as in a recurrent network), or each core uses separate weights (as in a feedforward network).  The non-$\lambda$, non-$(r,\gamma)$ variants of the predictron then correspond to standard (convolutional) feedforward and (unrolled) recurrent neural networks respectively. 

The third dimension is whether we include skip connections.  This is equivalent to defining the model step to output a change to the current state, $\Delta \s$, and then defining $\s^{k+1} = h(\s^k + \Delta \s^k)$, where $h$ is the non-linear function---in our case a ReLU, $h(x) = \mbox{max}(0, x)$. The deep network with skip connections is a variant of ResNet \citep{He:2015}.

Root mean squared prediction errors for each architecture are shown in Figure \ref{predictron_results_second_cube}. 
All $(r, \gamma, \lambda)$-predictrons (red lines) outperformed the corresponding feedforward or recurrent baselines (black lines) both in the random mazes and in pool.
We also investigated the effect of changing the depth of the networks (see appendix); the predictron outperformed the corresponding feedforward or recurrent baselines for all depths, with and without skip connections.

\subsection{Semi-supervised learning by consistency}

We now consider how to use the predictron for semi-supervised learning, training the model on a combination of labelled and unlabelled random mazes. Semi-supervised learning is important because a common bottleneck in applying machine learning in the real world is the difficulty of collecting labelled data, whereas often large quantities of unlabelled data exist.

\begin{figure*}[h]
\centering
\includegraphics[width=0.65\textwidth]{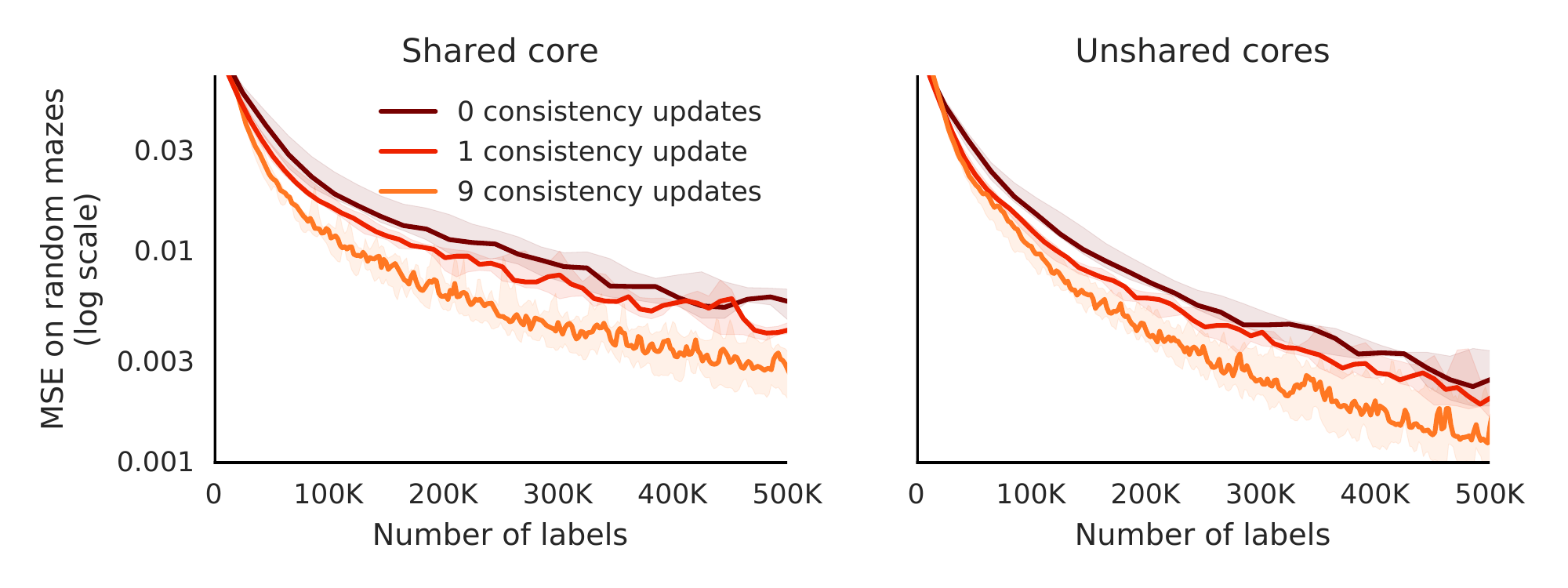}
\vspace{-1em}
\caption{\textbf{Semi-supervised learning.} Prediction errors of the $(r, \gamma, \lambda)$-predictrons (shared core, no skips) using 0, 1, or 9 consistency updates for every update with labelled data, plotted as function of the number of labels consumed. Learning performance improves with more consistency updates. \label{consistency_perf}}
\vspace{-0.2cm}
\end{figure*}

We trained a full $(r, \gamma, \lambda)$-predictron by alternating standard supervised updates with consistency updates, obtained by stochastically minimizing the consistency loss \eqref{eqn:consistency}, on additional unlabelled samples drawn from the same distribution. For each supervised update we apply either 0, 1, or 9 consistency updates.  Figure \ref{consistency_perf} shows that the performance improved monotonically with the number of consistency updates, measured as a function of the number of labelled samples consumed.

\subsection{Analysis of adaptive depth}

\begin{figure*}
    \centering
    \includegraphics[width=\textwidth]{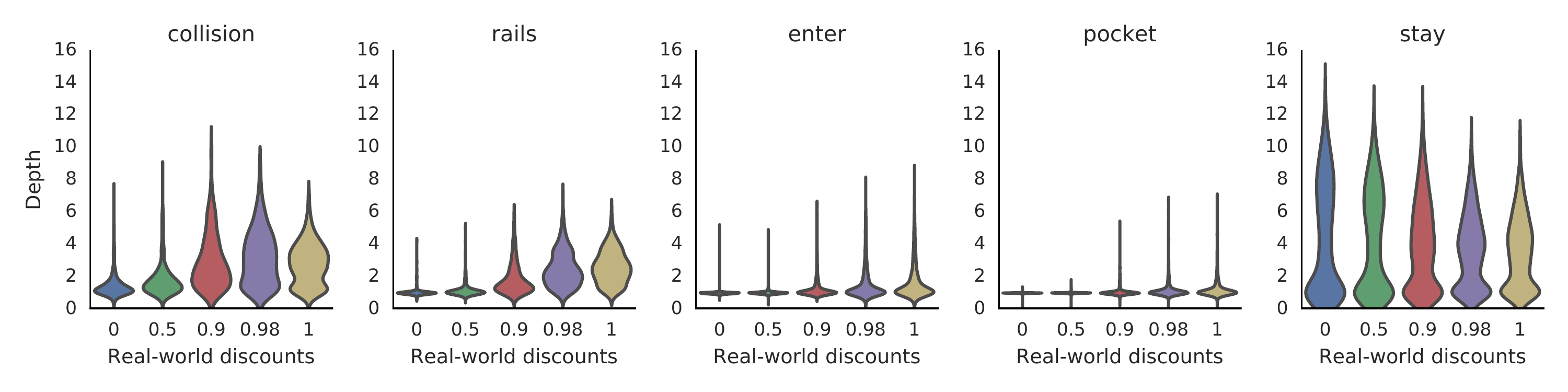}
    \vspace{-1em}
    \caption{\textbf{Thinking depth.} Distributions of thinking depth on pool for different types of predictions and for different real-world discounts. 
    }
    \label{fig:pool_violins}
\end{figure*}

In principle, the predictron can adapt its depth to `think more' about some predictions than others, perhaps depending on the complexity of the underlying target. We saw indications of this in Figure \ref{predictron_planning}.
We investigate this further by looking at qualitatively different prediction types in pool: ball collisions, rail collisions, pocketing balls, and entering or staying in quadrants. For each prediction type we consider several different time-spans (determined by the real-world discount factors associated with each pseudo-reward). Figure \ref{fig:pool_violins} shows distributions of \emph{depth} for each type of prediction.  The `depth' of a predictron is here defined as the effective number of model steps.  If the predictron relies fully on the very first value (i.e., $\l^0=0$), this counts as 0 steps.  If, instead, it learns to place equal weight on all rewards and on the final value, this counts as 16 steps.  Concretely, the depth ${\bm d}$ can be defined recursively as ${\bm d} = {\bm d}^0$ where
$
{\bm d}^k = \l^k( 1 + \d^k {\bm d}^{k+1})
$
and ${\bm d}^K = {\bm 0}$. Note that even for the same input state, each prediction has a separate depth.

The depth distributions exhibit three properties. First, different types of predictions used different depths.  Second, depth was correlated with the real-world discount for the first four prediction types. Third, the distributions are not strongly peaked, which implies that the depth can differ per input even for a single real-world discount and prediction type.
In a control experiment (not shown) we used a scalar $\lambda$ shared among all predictions, which reduced performance in all scenarios, indicating that the heterogeneous depth is a valuable form of flexibility.

\subsection{Using predictions to make decisions}

We test the quality of the predictions in the pool domain to evaluate whether they are well-suited to making  decisions.
For each sampled pool position,  we consider a set $I$ of different initial conditions (different angles and velocity of the white ball), and ask which is more likely to lead to pocketing coloured balls.
For each initial condition $s \in I$, we apply the $(r, \gamma, \lambda)$-predictron (shared cores, 16 model steps, no skip connections) to obtain predictions $\g^{\lambda}$. We ensemble the predictions associated to pocketing any ball (except the white one) with discounts $\gamma=0.98$ and $\gamma=1$. We select the condition $s^*$ that maximises this sum.

We then roll forward the pool simulator from $s^*$ and log the number of pocketing events. Figure~\ref{fig:domains} shows a sampled rollout, using the predictron to pick $s^*$. When providing the choice of $128$ angles and two velocities for initial conditions ($|I|=256$), this procedure resulted in pocketing 27 coloured balls in 50 episodes.  Using the same procedure with an equally deep convolutional network only resulted in $10$ pocketing events. 
These results suggest that the lower loss of the learned $(r, \gamma, \lambda)$-predictron translated into meaningful improvements when informing decisions. 
A video of the rollouts selected by the predictron is available at the following url: \url{https://youtu.be/BeaLdaN2C3Q}.

\section{Related work}
\citet{Lee:2015} introduced a neural network architecture where classifications branch off intermediate hidden layers. An important difference with respect to the $\lambda$-predictron is that the weights are hand-tuned as hyper-parameters, whereas in the predictron the $\lambda$ weights are learnt and, more importantly, conditional on the input. Another difference is that the loss on the auxiliary classifications is used to speed up learning, but the classifications themselves are not combined into an aggregate prediction; the output of the model itself is the deepest prediction.

\citet{graves:act} introduced an architecture with adaptive computation time (ACT), with a discrete (but differentiable) decision on when to halt, and aggregating the outputs at each pondering step. This is related to our $\l$ weights, but obtains depth in a different way; one notable difference is that the $\lambda$-predictron can use different pondering depths for each of its predictions.

Value iteration networks (VINs) \citep{Tamar:2016} also learn value functions end-to-end using an internal model, similar to the (non-$\lambda$) predictron. However, VINs plan via convolutional operations over the full input state space; whereas the predictron plans via imagined trajectories through an abstract state space. This may allow the predictron architecture to scale much more effectively in domains that do not have a natural two-dimensional encoding of the state space.

The notion of learning about many predictions of the future relates to work on predictive state representations~\citep[PSRs;][]{psr}, general value functions~\citep[GVFs;][]{Sutton:2011}, and nexting \citep{nexting}. Such predictions have been shown to be useful as representations~\citep{forecasts} and for transfer~\citep{uvfa}. So far, however, none of these have been considered for learning abstract models.

\citet{Schmidhuber:2015} discusses learning abstract models, but maintains separate losses for the model and a controller, and suggests training the model unsupervised to compactly encode the entire history of observations, through predictive coding. The predictron's abstract model is instead trained end-to-end to obtain accurate values.

\section{Conclusion}

The predictron is a single differentiable architecture that rolls forward an internal model to estimate external values. This internal model may be given both the structure and the semantics of traditional reinforcement learning models. But, unlike most approaches to model-based reinforcement learning, the model is fully abstract: it need not correspond to the real environment in any human understandable fashion, so long as its rolled-forward ``plans" accurately predict outcomes in the true environment.

The predictron may be viewed as a novel network architecture that incorporates several separable ideas. First, the predictron outputs a value by accumulating rewards over a series of internal planning steps. Second, each forward pass of the predictron outputs values at multiple planning depths. Third, these values may be combined together, also within a single forward pass, to output an overall ensemble value. Finally, the different values output by the predictron may be encouraged to be self-consistent with each other, to provide an additional signal during learning. Our experiments demonstrate that these differences result in more accurate predictions of value, in reinforcement learning environments, than more conventional network architectures. 

We have focused on value prediction tasks in uncontrolled environments. However, these ideas may transfer to the control setting, for example by using the predictron as a Q-network \citep{Mnih:2015}. Even more intriguing is the possibility of learning an internal MDP with abstract internal actions, rather than the MRP considered in this paper. We aim to explore these ideas in future work.

\newpage
\small
\bibliographystyle{icml2017}
\bibliography{paper}

 \newpage
 \newpage

 \appendix

\section{Architecture}
\label{apx:architecture}

The state representation $f$ is a two-layer convolutional neural network \citep{Lecun:1998}.
There is a \emph{core} $c$, again based on convolutions, that combines both MRP model and $\lambda$-network into a single repeatable module, such that $\s^{k+1}, \r^{k+1}, \d^{k+1}, \l^k = c(\s^k)$. This core is deterministic, and is duplicated $K$ times in the predictron with shared weights. (The predictron with unshared weights has $K$ distinct cores.) Finally, the value network $v$ is a fully connected neural network that computes $\v^k = v(\s^k)$.

\begin{figure}
\centering
\includegraphics[width=0.2\textwidth]{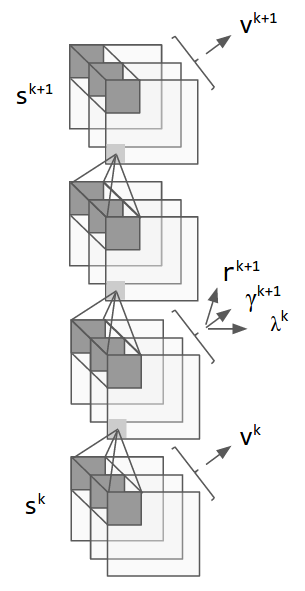}
\caption{The predictron core used in our experiments.}
\label{fig:core}
\end{figure}

Concretely, the $core$ (Figure \ref{fig:core}) consists first of a convolutional layer that maps into an intermediate (hidden) layer. From this layer, another two convolutions compute the next abstract state of the predictron.  Additionally, this same hidden layer is flattened and fed into three separate networks, with two fully connected layers each.  The outputs of these three networks represent the internal rewards, discounts, and lambdas.  A similar small network also hangs off the internal states, in addition to the core, and computes the values.
All convolutions use 3$\times$3 filters and a stride of one, and use padding to retain the size of the feature maps.  All feature maps have 32 channels.
The hidden layers within the MLPs have 32 hidden units.

In Figure \ref{fig:core} the convolutional layers are schematically drawn with three channels, flattening is represented by curly brakets, while the arrows represent the small multi-layer perceptrons which compute values, rewards, discounts and lambdas.

 We allow up to 16 model steps in our experiments, resulting in 52-layer deep networks---two convolutional layers for the state representations, $3 \times 16 = 48$ convolutional layers for the core steps, and two fully-connected layers for the values on top of the final state.  Between each two layers we apply batch normalization \citep{Ioffe:2015} followed by a ReLU non-linearity \citep{Glorot:2011}.  The value and reward networks end with a linear layer, whereas the discount and $\lambda$-networks additionally add a sigmoid non-linearity to ensure that these quantities are in $[0,1]$.
 
 For the illustrative maze experiment in Section \ref{sec:seqplans}, a smaller network architecture is employed with 6 model steps and convolutional feature maps of 16 channels. Additionally, the subnetworks to compute values, rewards, discounts, and lambdas are composed of a $1\times 1$ convolution with a stride of $1$ and 8 channels before a fully connected hidden layer of size 128. The rest of network architecture is as described above.

\section{Training}
\label{apx:training}

All experiments used the supervised (Monte-Carlo) update described in Section \ref{sec:updates} except for the semi-supervised experiment which used the consistency update described in Section \ref{sec:semi}. We update all parameters by applying the Adam optimiser \citep{Kingma:2015} to stochastic gradients of the corresponding loss functions. Each return is normalised by dividing it by its standard deviation (as measured, prior to the experiment, on a set of 20,000 episodes). In all experiments, the learning rate was 0.001, and the other parameters of the Adam optimiser were $\beta_1=0.9$, $\beta_2=0.999$, and $\epsilon=10^{-8}$. We used mini-batches of 100 samples.

\section{Comparing architectures of different depths}
\label{apx:depths}

\begin{figure*}
\centering
\includegraphics[width=0.8\textwidth]{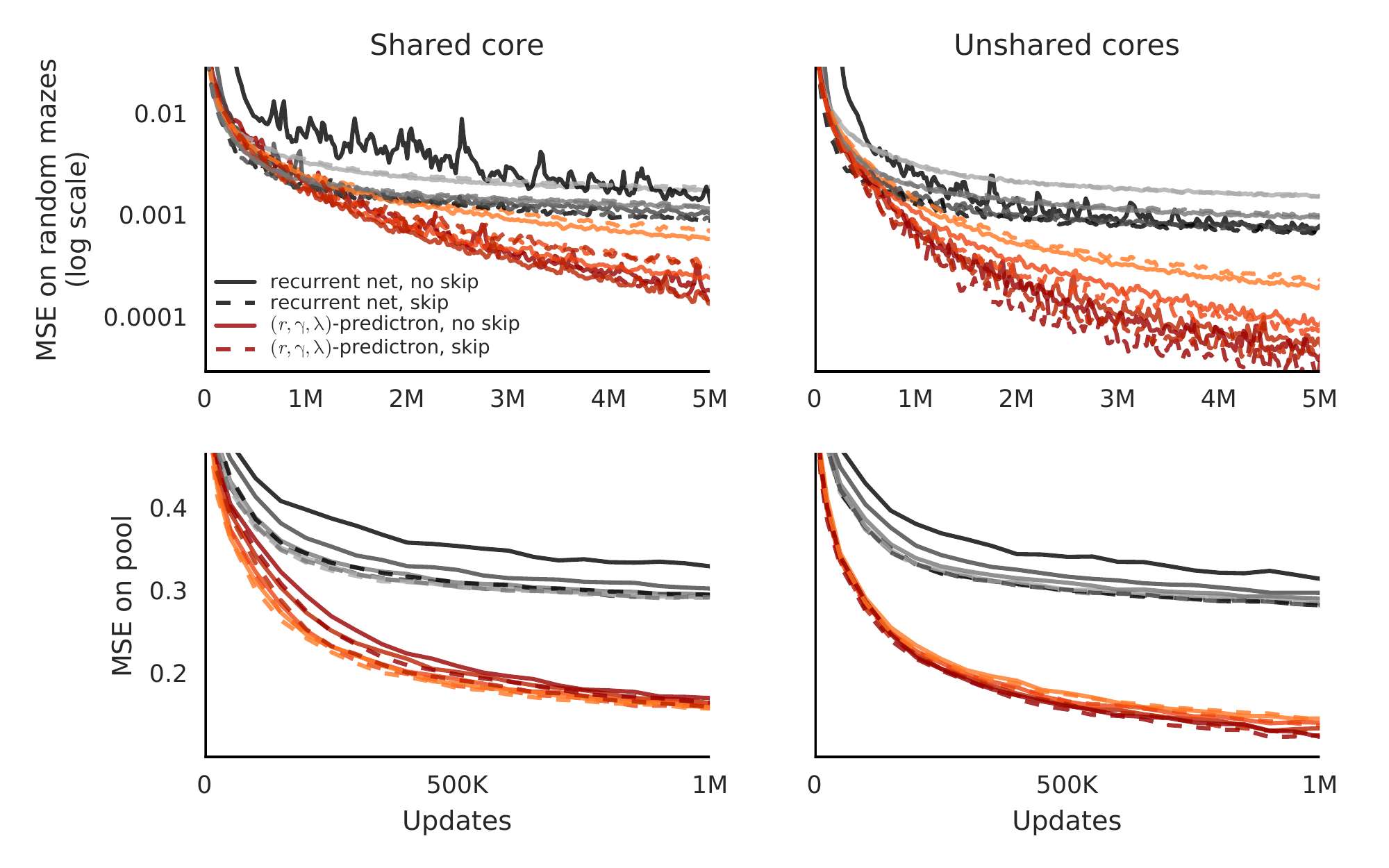}
\caption{\textbf{Comparing depths.} Comparing the $(r, \gamma, \lambda)$-predictron (\red{\textbf{red}}) against more conventional deep networks (\textbf{black}) for various depths (2, 4, 8, or 16 model steps, corresponding to 10, 16, 28, or 52 total layers of depth). Lighter colours correspond to shallower networks.  Dashed lines correspond to networks with skip connections.}
\label{fig:depth}
\end{figure*}

We investigated the effect of changing the depth of the networks, with and without skip connections.  Figure \ref{fig:depth} in shows that skip connections (dashed lines) make the conventional architectures (black/grey lines) more robust to the depth (i.e., the black/grey dashed lines almost overlap, especially on pool), and that the predictron outperforms the corresponding feedforward or recurrent baselines for all depths, with and without skips.

\begin{figure}
\centering
\includegraphics[width=0.25\textwidth]{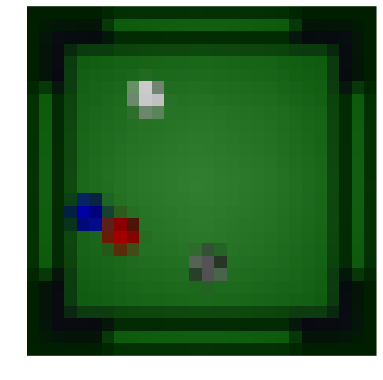}
\caption{\textbf{Pool input frame.} An example of a 28x28 RGB input frame in the pool domain.}
\label{fig:pool28}
\end{figure}

\section{Capacity comparisons}
In this section, we present some additional experiments comparing the predictron to more conventional deep networks.  The purposes of these experiments are 1) to show that the conclusions obtained above do not depend on the precise architecture used, and 2) to show that the structure of the network---whether we use a predictron or not---is more important than the raw number of parameters.

Specifically, we again consider the same 20 by 20 random mazes, and the pool task described in the main text.  As described in Section \ref{apx:architecture}, for the results in the paper we used an encoder that preserved the size of the input plans, $20\times 20$ for the mazes and $28\times 28$ for pool.  Each convolution had 32 channels and therefore the abstract states were $20\times 20 \times 32$ for the mazes and $28\times 28 \times 32$ for pool.

We now consider a different architecture, where we no longer pad the convolutions used in the encoder.  For the mazes, we still use two layers of $3\times 3$ stride-$1$ convolutions, which means the planes reduce in size to $16\times 16$.  This means that the abstract states are about one third smaller.  For pool, we use three $5 \times 5$ stride-$1$ convolutions, which bring us from $28 \times 28$ down to $16 \times 16$ as well.  So, the abstract states are now of equal size for both experiments.  For pool, this is approximately a two-thirds reduction, which helps reduce the compute needed to run the model.

Most of the parameters in the predictron are in the fully connected layers.  Previously, the first fully connected layer for each of the internal values, rewards, discounts, and $\lambda$-parameters would take a flattened abstract state, and then go into 32 hidden nodes.  This means the number of parameters in this layer were $20\times 20\times 32 \times 32 = 409,600$ for the mazes and $28 \times 28 \times 32 \times 32 = 802,816$ for pool.  The predictron with shared core would have four of these layers, one for each of the internal values, rewards, discounts, and $\lambda$s, compared to one for the deep network which only has values.  We change this in two ways.  First, we add a $1\times 1$ convolution with a stride of $1$ and 8 channels before the first fully connected layer for each of these outputs. This reduces the number of channels, and therefore the number of parameters in the subsequent fully-connected layer, by one fourth.  Second, we tested three different numbers of hidden nodes: 32, 128, or 512.

\begin{figure*}
\centering
\includegraphics[width=0.95\textwidth]{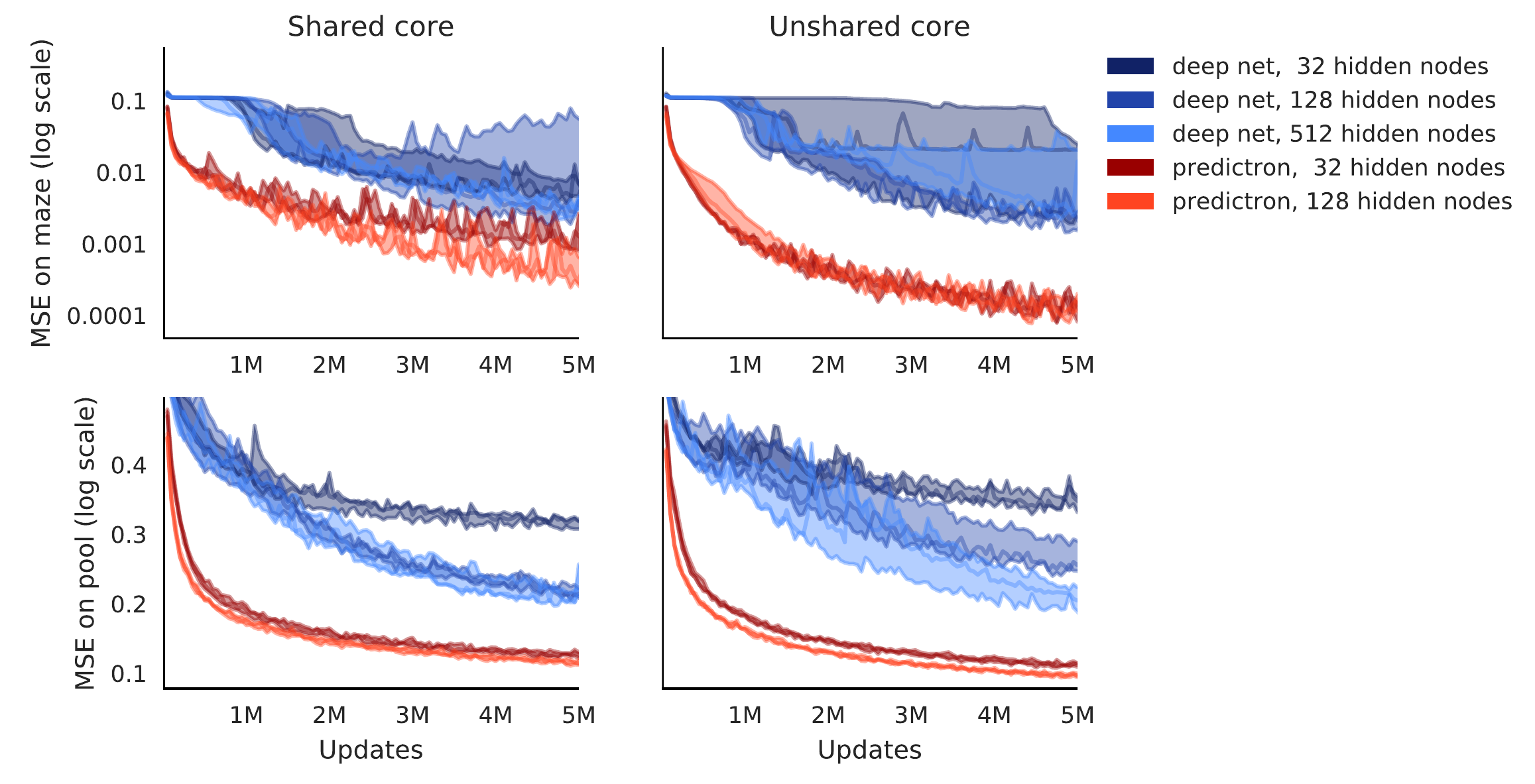}
\caption{\textbf{Comparing depths.} Comparing the $(r, \gamma, \lambda)$-predictron (\red{\textbf{red}}) against more conventional deep networks (\blue{\textbf{blue}}) for different numbers hidden nodes in the fully connected layers, and therefore different total numbers of parameters.  The deep networks with 32, 128, and 512 nodes respectively have 381,416, 1,275,752, and 4,853,096 parameters in total.  The predictrons with 32 and 128 nodes respectively have 1,275,752, and 4,853,096 parameters in total.  Note that the number of parameters for the 32 and 128 node predictrons are exactly equal to the number of parameters for the 128 and 512 node deep networks.}
\label{fig:capacity}
\end{figure*}

The deep network with 128 hidden nodes for its values has the exact same number of parameters as the $(r, \gamma, \lambda)$-predictron with 32 hidden nodes for each of its outputs.  Before, the deep network had fewer parameters, because we kept this number fixed at 32 across experiments.  This opens the question of whether the improved performance of the predictron was not just an artifact of having more parameters.  We tested this hypothesis, and the results are shown in Figure \ref{fig:capacity}.

Figure \ref{fig:capacity} shows that in each setting---on the mazes and pool, and with or without shared cores---both. The predictrons always performed better than all the deep networks.  This includes the 32 node predictron (darkest red) compared to the 512 node deep network (lightest blue), even though the latter has approximately 4 times as many parameters (1.27M vs 4.85M).  This means that the number of parameters mattered less than whether or not we use a predictron.

\section{Additional domain details}
\label{apx:domains}

We now provide some additional details of domains.

\subsection{Pool}

To generate sequences in the Pool domain, the initial locations of 4 balls of different colours are sampled at random. The white ball is the only one moving initially. Its velocity has a norm sampled uniformly between 7 and 14. The initial angle is sampled uniformly in the range (0, 2$\pi$). From the initial condition, the Mujoco simulation is run forward until all balls have stopped moving; sequences that last more than 151 frames are rejected, and a new one is generated as replacement. Each frame is rendered by Mujoco as a 280x280 RGB image, and subsequently downsampled through bilinear interpolation to a 28x28 RGB input (see Figure~\ref{fig:pool28} for an example). Since the 280 signals described in Section 6.1 as targets for the Pool experiments have very different levels of sparsity, resulting in values with very different scales, we have normalised the pseudo returns. The normalization procedure consisted in dividing all targets by their standard deviation, as empirically measured across an initial set of 20,000 sequences.

\subsection{Random Mazes}

\subsubsection{First Task}

The mazes are generated by ensuring that around 15\% of locations are walls. The policy takes as observation the wall configuration in four locations adjacent to its position and maps each of these configurations to an action. For each maze, the policy is stepped for 60 steps from a uniformly random start location. The target $\g$ indicates whether the trajectory has traversed each maze location.  

\subsubsection{Second Task}

To generate mazes we first determine, with a stochastic line search, a number of walls so that the top-left corner is connected to the bottom-right corner (both always forced to be empty) in approximately 50\% of the mazes. We then shuffle the walls uniformly randomly.  For 20 by 20 mazes this means 70\% of locations are empty and 30\% contain walls.  More than a googol different such 20-by-20 mazes exist (as ${398 \choose 120} >10^{100}$).

\end{document}